\documentclass{ifacconf}

\usepackage{graphicx}
\usepackage{natbib}        
\usepackage{amsmath,amssymb,amsfonts}
\usepackage{xcolor}

\newcommand{\tr}{^\mathrm{T}}
\newcommand{\ntr}{^\mathrm{-T}}

\newcommand{\abs}[1]{\left\lvert#1\right\rvert}
\newcommand{\norm}[1]{\left\lVert#1\right\rVert}
\newcommand{\defeq}{\doteq}
\newcommand{\confreg}[1][2]{\mathcal{C}_{#1}}
\newcommand{\outappr}[1][2]{\mathcal{O}_{#1}}

\DeclareMathOperator*{\argmin}{arg\,min}

\newcommand{\BP}{\mathbb{P}}

\newcommand{\BR}{\mathbb{R}}

\allowdisplaybreaks

\begin{document}
\begin{frontmatter}

\title{Distribution-Free Confidence Ellipsoids for Ridge Regression with PAC Bounds\thanksref{footnoteinfo}}

\thanks[footnoteinfo]{
This research was supported by the European Union within the framework of the National Laboratory for Autonomous Systems, RRF-2.3.1-21-2022-00002; the work was also partially supported by the ``Robust Uncertainty Quantification for Learning and Control'' research project of the National Research, Development and Innovation Office of Hungary (NKFIH), grant number 153390.}

\author[First]{Szabolcs Szentpéteri} 
\author[First,Second]{Balázs Csanád Csáji} 

\address[First]{Institute for Computer Science and Control (SZTAKI), Hungarian Research            Network (HUN-REN), 13-17 Kende utca, H-1111, Budapest, Hungary (e-mail: szabolcs.szentpeteri@sztaki.hu).}
\address[Second]{Department of Probability Theory and Statistics, Institute of Mathematics, E\"otv\"os Lor\'and University (ELTE),
		1/C Pázmány Péter sétány, H-1117, Budapest, Hungary (e-mail: csaji@sztaki.hu)}
 
\begin{abstract}
Linearly parametrized models are widely used in control and signal processing, with the least-squares (LS) estimate being the archetypical solution. When the input is insufficiently exciting, the LS problem may be unsolvable or numerically unstable. This issue can be resolved through regularization, typically with ridge regression. Although regularized estimators reduce the variance error, it remains important to quantify their estimation uncertainty. A possible approach for linear regression is to construct confidence ellipsoids with the Sign-Perturbed Sums (SPS) ellipsoidal outer approximation (EOA) algorithm. The SPS EOA builds non-asymptotic confidence ellipsoids under the assumption that the noises are independent and symmetric about zero. This paper introduces an extension of the SPS EOA algorithm to ridge regression, and derives probably approximately correct (PAC) upper bounds for the resulting region sizes. Compared with previous analyses, our result explicitly show how the regularization parameter affects the region sizes, and provide tighter bounds under weaker excitation assumptions. Finally, the practical effect of regularization is also demonstrated via simulation experiments.
\end{abstract}


\end{frontmatter}

\vspace*{-5mm}
\section{Introduction}
Constructing mathematical models of dynamical systems from observed data is the main concern of system identification \citep{Ljung1999}, however, the problem also arise in other fields such as machine learning and statistics. Stochastic linear models remain among the most widely used due to their simplicity and theoretical guarantees. A well-know example is linear regression (LR), where the least-squares (LS) estimator is the most typically applied.

A drawback of LS is that its susceptible to numerical instability. To address this issue, several regularized estimators have been proposed, such as ridge regression (RR). One of the core techniques of regularization is to shrink the estimate a bit, in order to obtain smaller variance. Applying regularized methods has a rich history in system identification, however, as with LS, the classical result are primarily asymptotic in nature \citep{pillonetto2022}.

In recent years, a paradigm shift has taken place, marked by increasing attention to non-asymptotic guarantees.
The finite-sample behavior of (regularized) estimators are now extensively studies with the goal of providing strong mathematical guarantees for the estimates that hold for any finite sample size. Such results are crucial if strong stability or robustness requirements must be satisfied.

While the sample complexity of the ordinary LS estimator has been studied for observable \citep{simchowitz18a}, non-observable \citep{oymak2021}, and even unstable \citep{faradonbeh2018} linear stochastic state-space models, as well as for finite impulse response (FIR) models \citep{Djehiche2021}, the analysis of other LS-based estimators remains limited.
A finite-sample error bound for a regularized estimator in the non-observable state-space setting is presented in \citep{Sun2022}, where the authors apply a Hankel nuclear norm regularization to identify a low-order system.
A different approach, which also estimates the Markov parameters of the system, is proposed in \citep{He2024}, which uses the weighted least-squares method and provides non-asymptotic guarantees for the resulting estimation error.
For observable state-space models, a regularized estimator based on the Schatten $p$-norm together with a sample complexity analysis is presented in \citep{park2025}.

Besides concentration inequalities and probably approximately correct (PAC)-based bounds, another way to quantify the uncertainty of an estimate is to build confidence regions. An algorithm that generates non-asymptotic confidence regions and ellipsoids is the Sign-Perturbed Sums (SPS) method \citep{Csaji2015}. A significant advantage of SPS over PAC-type bounds is that it is data-driven, i.e., it requires no hyper-parameters to construct an ellipsoid which contains the true parameter with high probability. Moreover, it is distribution-free and therefore does not rely on any prior knowledge of the noises distributions. In contrast, all of the PAC-bound-based results referenced above assume Gaussian noise or subgaussian tail distributions.

The original SPS method is an indicator function, which decides whether an input parameter is included in the region. To give a compact representation of the region that can be computed efficiently without parameter-by-parameter evaluation, an ellipsoidal outer approximation (EOA) of SPS was proposed by \cite{Csaji2015} for linear regression. A non-asymptotic bound on the size of SPS EOA regions for linear regression is proven in \citep{szentpeteri2025finite}, which also shows that these regions shrink at an optimal rate. Possible extensions of the SPS indicator function to regularized estimates including ridge regression and LASSO were presented in \citep{csaji2019}.

In this work we propose an SPS EOA algorithm for ridge regression and analyze the sample complexity of the obtained confidence ellipsoids. Our main contributions are:
\begin{enumerate}
    \item An ellipsoidal outer approximation algorithm of SPS for ridge regression (RR-SPS EOA) is introduced.
    \item PAC upper bounds are derived for the size of RR-SPS EOA confidence regions,
    which explicitly characterize how the regularization affects the magnitude.
    %Demonstrating that using a 
    %different 
    \item It is shown that using a
    Mahalanobis metric, instead of the Euclidean one,
    leads to relaxed assumptions.
    \item An
    %Improving upon an 
    intermediate result of \citep{szentpeteri2025finite} is improved to obtain tighter PAC bounds.
    \item Simulation results are provided to illustrate the effect of regularization and to compare the empirical and theoretical sizes of RR-SPS EOA confidence sets.
\end{enumerate}
{{\em Notations:} throughout the paper we will use the following notations: $[n]\defeq\{1,\dots,n\}$, $\bar{A}\defeq\tfrac{1}{n}A$. The eigenvalues of $A$ are denoted by $\lambda_i(A)$, while $\lambda_{\text{min}}(A)$ and $\lambda_{\text{max}}(A)$ are the minimum and maximum eigenvalues of $A$. $\|x\|$ stands for the Euclidean norm, and $\|x\|^2_A\defeq x\tr Ax$. We use $I_d \in \BR^{d\times d}$ for the identity matrix, and $\text{diag}(d_1,\dots,d_d)=\text{diag}(d_i)_{i=1}^d$ for a diagonal matrix, defined by its diagonal elements.}

\section{Problem setting}

Consider the following linear regression problem
\begin{equation}\label{equ:system}
    Y_t \,\defeq\, \varphi_t\tr \theta^* + W_t,
\end{equation}
for $t \in [n]$, where $Y_t$ is the scalar output, $\varphi_t \in \BR^d$ is a deterministic regressor, $\theta^*\in \BR^d$ is the true parameter and $W_t$ is the scalar random noise. We are given a sample of $n$ input vectors: $\varphi_1, \dots, \varphi_n$ and noisy outputs: $Y_1, \dots, Y_n$.

We introduce the following vectors:
\begin{align}
    \tilde{\Phi} &\defeq \begin{bmatrix}
        \varphi_1
        \varphi_2
        \dots
        \varphi_n
    \end{bmatrix}\tr,\quad
    \tilde{w} \defeq \begin{bmatrix}
        W_1
        W_2
        \dots
        W_n
    \end{bmatrix}\tr,\notag\\
    \tilde{y} &\defeq \begin{bmatrix}
        Y_1
        Y_2
        \dots
        Y_n
    \end{bmatrix}\tr.
\end{align}
We aim to construct confidence regions around the ridge regression (RR) estimate, defined by
\begin{equation}
    \hat{\theta} \defeq \argmin_{\theta \in \BR^d} \tfrac{1}{2}\|\tilde{y}-\tilde{\Phi}\theta\|^2 + \tfrac{\lambda}{2}\|\theta\|^2,
\end{equation}
where $\lambda > 0$ is a hyper-parameter. It is well-know that RR can be formulated as a least-squares problem, i.e., $\argmin_{\theta \in \BR^d} \tfrac{1}{2}\|y-\Phi\theta\|^2$, with the following extensions:
\begin{align}\label{equ:extended_matrices}
    &\Phi \defeq \begin{bmatrix}
        \tilde{\Phi} \\
        \sqrt{\lambda}I
        % \sqrt{\tfrac{\lambda}{2}}I
    \end{bmatrix},\qquad
    w\defeq \begin{bmatrix}
        \tilde{w}\\
        -\sqrt{\lambda}\theta^*
        % -\sqrt{\tfrac{\lambda}{2}}\theta^*
    \end{bmatrix},\qquad
    y \defeq \begin{bmatrix}
        \tilde{y}\\
        0
    \end{bmatrix},
\end{align}
and the solution is famously $\hat{\theta} = (\tilde{\Phi}\tr\tilde{\Phi} + \lambda I_d)^{-1}\tilde{\Phi}\tr \tilde{y}$.

Then, the (averaged) non-extended and extended sample covariance matrices and the RR estimate $\hat{\theta}$ are
\begin{align}
    \tilde{R} &\defeq \tilde{\Phi}\tr  \tilde{\Phi},
    &R &\defeq \Phi\tr  \Phi = \tilde{\Phi}\tr  \tilde{\Phi} + \lambda I_d = \tilde{R}+ \lambda I_d, \notag\\
    \bar{R} &\defeq \tfrac{1}{n}R,
    &\hat{\theta} &= R^{-1}\Phi\tr y.
\end{align}

\subsection{Assumptions}
Our first two assumptions are the same as in the sample complexity analysis of SPS EOA regions for the least-squares estimate \citep{szentpeteri2025finite}. As we prove PAC upper bounds on the size of RR-SPS ellipsoids using a Mahalanobis metric, instead of the Euclidean, we were able to relax the excitation assumption (A3) from our previous work. Another difference is that the coherence assumption on the input (Assumption \ref{assu:tilde_Phi_U_coherence}) uses the SVD-decomposition and not the QR-decomposition. Due to regularization, we also make an additional assumption (Assumption \ref{assu:true_param}) that was not needed for linear regression.
\begin{assum}\label{assu:noise}
{\em The noise sequence $\{W_t\}$ is independent and contains nonatomic, $\sigma$-subgaussian random variables whose probability distributions are symmetric about zero. }
\end{assum}
A random variable $W$ is $\sigma${\em-subgaussian} \cite[Definition 2.2]{wainwright_2019}, if 
for all $\lambda \in \mathbb{R},$ there is a 
$\sigma>0$, such that
\begin{align}
\mathbb{E}\big[\exp(\lambda (W\!-\mathbb{E}[W]))\big]\, \leq\, \exp\!{\left(\frac{\lambda^2\sigma^2}{2}\right)}.
\end{align}
The $\sigma$-subgaussianity of the noise sequence is only assumed to provide non-asymptotic upper bounds on the sizes of the RR-SPS ellipsoids, the algorithm itself can build confidence ellipsoids for independent and symmetric noises. Nonetheless, subgaussianity covers a wide range of possible distributions, and allows even non-stationary sequences.
\begin{assum}\label{assu:R_nonsigular}
{\em Any $d$ regressors span the whole space, $\BR^d$. }
\end{assum}
This assumption ensures that the RR-SPS ellipsoids are bounded under suitable assumptions on the perturbations.
{\begin{assum}\label{assu:tilde_Phi_U_coherence}
{\em Let\, $\tilde{\Phi} = U_{\tilde{\Phi}}\Sigma_{\tilde{\Phi}}V_{\tilde{\Phi}}\tr $ be the SVD decomposition of\, $\tilde{\Phi}$. 
There are constants $\kappa >0$ and\, $0 <\rho \leq 1$, such that the following upper bound holds {for all\, $n > d$}:
\begin{align}
    \mu(\tilde{\Phi})\, \defeq\, \frac{n}{d} \max_{1\leq i \leq n}\norm{U_{\tilde{\Phi}}\tr e_i}^2 \leq\, \kappa\, n^{1-\rho},
\end{align}
 where $\mu(\tilde{\Phi})$ is called the coherence of\, $\tilde{\Phi}$.}
\end{assum}}
The coherence \citep[Definition 1.2]{Candes2008ExactMC} assumption guarantees that the input is not too ``unevenly distributed'' among the regressors. Note that this assumption is almost surely satisfied if the excitations are i.i.d. random vectors with a positive definite covariance matrix, which is a very common assumption in control theory.
\begin{assum}\label{assu:true_param}
{\em    The true parameter vector $\theta^*$ is included in an Euclidean norm ball with radius $\ell > 0$, hence,
    \begin{equation}
        \norm{\theta^*} \leq \ell.
    \end{equation}}
\end{assum}
This last assumption is needed since the size of the RR-SPS EOA regions depend on $\theta^*$ in case of ridge regression.
\section{The ridge regression SPS algorithm}
\subsection{The ridge regression SPS-Indicator algorithm}
A possible generalization of SPS for ridge regression was proposed in \citep{csaji2019}. Here, we present a different solution, which applies $\bar{R}$ as a correction term instead of ${R}^{-1}\tilde{R}{R}^{-1}$. This way, the shape of the obtained region matches the shape of the asymptotic confidence region around the least-squares estimate. Furthermore, it allows a simplified ellipsoidal outer approximation construction. The proposed RR-SPS algorithm consists of two parts, the initialization, which sets the required global parameters and generates the random signs, and the indicator function, which decides whether a given $\theta$ is included in the region.
The initialization and indicator function is given in Table \ref{tab:sps_init} and \ref{tab:sps_indicator}. Using these algorithms, the $p$-level RR-SPS confidence region can be defined as
\vspace{1mm}
\begin{equation}
        \confreg[n]\, \defeq\, \{\,\theta \in \BR^d\text{ : RR-SPS-Indicator}(\theta) = 1\,\}.
\vspace{1mm}        
\end{equation}
The constructed confidence region $\confreg[n]$ has exact coverage probability, therefore, the following theorem holds:
\begin{thm}\label{thm:exact_confidence}
    {\em
    Assuming the noise sequence $\{W_t\}$ contains independent random variables that are distributed symmetrically about zero, the coverage probability of the constructed confidence region is exactly $p$, that is,
    \vspace{1mm}
    \begin{equation}
            \BP(\theta^* \in \confreg[n]) = 1-\frac{q}{m} = p.
            \vspace{1mm}
    \end{equation}
    }
\end{thm}
\begin{pf}
    The exact coverage of RR-SPS indicator regions can be proven similarly to the case of SPS indicator regions in \citep{Csaji2015}. The main idea is to show that $\{ \| S_i(\theta^*)\|^2\}_{i=0}^{m-1}$ are exchangeable. Note that regularization does not affect exchangeability, since $\Phi\tr\varepsilon(\theta^*)=\tilde{\Phi}\tr(\tilde{y}-\tilde{\Phi}\theta^*) - \lambda\theta^*$ and $\forall i : \Phi\tr D_i\varepsilon(\theta^*) =\tilde{\Phi}\tr\tilde{D}_i(\tilde{y}-\tilde{\Phi}\theta^*) - \lambda\theta^* $, and the exchangeability of $\{ \| \tilde{\Phi}\tr\tilde{D}_i(\tilde{y}-\tilde{\Phi}\theta^*)\|\}_{i=0}^{m-1}$, where $\tilde{D}_0=I_n$, were proven in the above referenced paper.
\end{pf}
\begin{table}[t]
    \centering
    \caption{Pseudocode: \hspace{-0.5mm}RR-SPS-Initialization\hspace{-0.5mm} $(\hspace{0.2mm}p\hspace{0.2mm})$}
    \vspace{-2.5mm}
    \label{tab:sps_init}
    \setlength{\tabcolsep}{3pt}
    \begin{tabular}{p{10pt}p{205pt}}
        \hline
        1.& 
        Given a (rational) confidence probability $p \in (0,1)$, set integers $m > q >0$ such that $p = 1 - q/m$. \\
        2.& 
        Generate $n(m-1)$ i.i.d random signs $\{\alpha_{i,t}\}$ with
        \[\mathbb{P}(\alpha_{i,t} = 1) = \, \mathbb{P}(\alpha_{i,t} = -1) = 0.5\] 
        for all integers $1 \leq i \leq m-1$ and $1 \leq t \leq  n$ and construct the following $(n+d)\times (n+d)$-sized diagonal matrices containing these random signs
        \begin{equation*}
            D_i \defeq \begin{bmatrix}
			\tilde{D}_i & 0\\
             0&I_d
		\end{bmatrix},
        \end{equation*}
        where $\tilde{D}_i\defeq\text{diag}(\alpha_{i,1},\dots,\alpha_{i,n})$.\\
		% \[D_i \, \defeq \, \begin{bmatrix}
		% 	\alpha_{i,1} & & &   \\
		% 	& \ddots & &  \\
		% 	& & \alpha_{i,n} &  \\
  %           & & & I_d
		% \end{bmatrix}\!.\vspace{-3mm}
		% \]\\
        3.&
        Generate a permutation $\pi$ of the set $\{0,\dots, m - 1\}$ randomly, where each of the $m!$ possible permutations has the same probability $1/(m!)$ to be selected.\\
        \hline
    \end{tabular}
    \vspace{3mm}
\end{table}
\begin{table}[t]
    \centering
    \caption{Pseudocode: RR-SPS-Indicator $(\hspace{0.2mm}\theta\hspace{0.2mm})$}
    \vspace{-2.5mm}		
    \label{tab:sps_indicator}
    \setlength{\tabcolsep}{3pt}
    \begin{tabular}{p{10pt}p{205pt}}
        \hline
        1.& 
        For a given $\theta$, compute the prediction error vector\vspace{-0.5mm}
        \[ \varepsilon(\theta) \defeq y - \Phi\theta\vspace{-4mm}.\]\\
        2.& 
        Evaluate\vspace{-1mm}
        \[ S_0(\theta) \defeq \tfrac{1}{n}\bar{R}^{-\frac{1}{2}}\Phi\tr\varepsilon(\theta), \hspace{2mm} \text{and} \hspace{2mm}
        S_i(\theta) \defeq \tfrac{1}{n}\bar{R}^{-\frac{1}{2}}\Phi\tr D_i\varepsilon(\theta),\vspace{-0.5mm} \]
        for all indices $1 \leq i \leq m-1$. \\
        3.& 
        Order scalars $\{\|S_i(\theta)\|^2\}$ according to $\succ_{\pi}$, where ``$\succ_{\pi}$'' is ``$>$'' with random tie-breaking \citep{Csaji2015}. \\
        4.& 
        Compute the rank 
        of $\norm{S_0(\theta)}^2$ among $\{\norm{S_i(\theta)}^2\}_{i=1}^{m-1}:$ \vspace{-1.5mm}
        \begin{equation*}
            \mathcal{R}(\theta) \defeq \Bigg[1+\sum_{i=1}^{m-1}\mathbb{I}\left(\|S_0(\theta)\|^2 (\theta)\succ_{\pi} \|S_i(\theta)\|^2\right)\Bigg].
            \vspace{-2mm}
        \end{equation*}\\
        5.&
        Return 1 if $\mathcal{R}(\theta) \leq m - q$, otherwise return 0.\\
        \hline
    \end{tabular}
    \vspace{1mm}
\end{table}
\subsection{The ridge regression SPS EOA}
The RR-SPS-Indicator($\theta$) algorithm (Table \ref{tab:sps_indicator}) can only test whether an input parameter is in the confidence region. In practice, a compact representation of the confidence set is often needed. Building a region by evaluating different parameters would be computationally ineffective, therefore, for linear regression problems, an ellipsoidal outer approximation (EOA) algorithm was introduced in \citep{Csaji2015}, where the center of the ellipsoid is the least-squares estimate. The radius of the ellipsoid can be obtained by solving convex semidefinite (SDP) programs.

We propose a generalization of this outer-ellipsoid for ridge regression. In this case, the center of the ellipsoid is the RR estimate and the radius is computed from the extended matrices \eqref{equ:extended_matrices}, however, the convex semidefinite programs have the same mathematical form. The difference lies in the definition of regressor matrix $\Phi$ and output vector $y$. This is a consequence of the applied correction term $\bar{R}$, which is the covariance matrix of the ellipsoid. This means that
the confidence 
region given by the RR-SPS EOA is 
\begin{equation}\label{equ:SPS_EOA_conf}
    \confreg[n]\, \subseteq\, \outappr[n]\, \defeq \bigl\{\hspace{0.3mm}\theta \in \mathbb{R}^d : (\theta - \hat{\theta})\tr  \bar{R} (\theta - \hat{\theta}) \leq r\hspace{0.3mm}\bigl\},
    \vspace{1mm}	
\end{equation}
where $r$ can be computed from the solutions of the following optimization problems for $i \in [\hspace{0.3mm}m-1\hspace{0.2mm}]$:
\begin{equation}\label{equ:sps_eoa_cvx}
\begin{aligned}
    \min \quad & \gamma\\
    \textrm{s.t.} \quad & \xi \geq 0 \\
                        & \begin{bmatrix}
                            -I + \xi A_{i} & \xi b_{i}\\
                            \xi b_{i}\tr  & \xi c_{i} + \gamma
                        \end{bmatrix} \succeq 0,
\end{aligned}
\vspace{1mm}
\end{equation}
where
\begin{align}
A_{i} &\defeq I - R^{-\frac{1}{2}}Q_{i}R^{-1}Q_{i}R^{-\frac{1}{2}},\\[1mm]
b_{i} &\defeq \tfrac{1}{\sqrt{n}}R^{-\frac{1}{2}}Q_{i}R^{-1}\left(\psi_{i} - Q_{i}\hat{\theta}\right),\notag\\[1mm]
c_{i} &\defeq\tfrac{1}{n}(-\psi_{i}\tr  R^{-1}\psi_{i} + 2\hat{\theta}\tr  Q_{i}R^{-1}\psi_{i}-\hat{\theta}\tr Q_{i}R^{-1} Q_{i}\hat{\theta}),\notag\\[-4mm]\notag
\end{align}
and $\psi_{i}$ and $Q_{i}$ are defined as
\begin{align}\label{equ:def_D}
    &Q_{i} \defeq \Phi\tr D_i\Phi,\qquad \psi_{i}\defeq \Phi\tr D_iy.
\end{align}
The detailed computation of $r$ and the RR-SPS EOA algorithm as a whole is presented in Table \ref{tab:sps_eoa}. Since $\outappr[n]$ is an outer approximation, it satisfies that
\vspace{1mm}
\begin{align}
    \BP(\theta^* \in \outappr[n]) \geq 1-\frac{q}{m} = p.
\end{align}
\begin{table}[t]
    \centering
    \caption{Pseudocode: RR-SPS EOA}
    \vspace{-1.5mm}		
    \label{tab:sps_eoa}
    \setlength{\tabcolsep}{3pt}
    \begin{tabular}{p{10pt}p{205pt}}
        \hline
        1.& 
        Compute the ridge regression estimate,\vspace{-0.5mm}
        \[ \hat{\theta} = R^{-1}\Phi\tr y.\vspace{-4mm}\]\\
        2.&
        For $i \in [\hspace{0.3mm}m-1\hspace{0.2mm}]$, solve the optimization problem \eqref{equ:sps_eoa_cvx}, and let $\gamma_i^*$ be the optimal value.\\
        3.& 
        Let $r$ be the $q$th largest $\gamma_i^*$ value. \\
        4.& 
        The outer approximation of the RR-SPS confidence
            region is given by the ellipsoid 
            \begin{equation*}
                    \outappr[n]\, \defeq \bigl\{\hspace{0.3mm}\theta \in \mathbb{R}^d : (\theta - \hat{\theta})\tr  {\bar{R}} (\theta - \hat{\theta}) \leq r\hspace{0.3mm}\bigl\}.\vspace{-2.5mm}
            \end{equation*}\\
        \hline
    \end{tabular}
    \vspace{2mm}
\end{table}
\section{Sample complexity of RR-SPS EOA}
In this section we present a sample complexity theorem for the RR-SPS EOA regions.
The theorem and its proof build upon the ideas of \citep[Theorem 3]{szentpeteri2025finite}, however, it $i)$ explicitly shows how regularization affects the size of the regions, $ii)$ uses the Mahalanobis metric, leading to relaxed assumptions, and $iii)$ improves an intermediate lemma, to provide a tighter bound.
\begin{thm}\label{thm:sample_complex_eoa}
    {\em
    Under Assumptions \ref{assu:noise}, \ref{assu:R_nonsigular}, \ref{assu:tilde_Phi_U_coherence}, and \ref{assu:true_param}, the following concentration inequality holds for the size of RR-SPS EOA confidence regions.
    For all\, $\delta \in (0, 1)$ and\, $n \geq \lceil g^{1/\rho}(\frac{\delta}{m-q})\rceil$ with probability at least\, $1-\delta,$ we have
\begin{align}
    &\sup_{\theta \in \outappr[n]}\|\theta - \hat{\theta}\|^2_{\bar{R}} \leq\\
    &\dfrac{\left(f\left(\frac{\delta}{m-q}\right)+2\lambda\ell^2d\right)\left(n^{\frac{\rho}{2}}\left(1+\frac{2\lambda}{\lambda_{\min}(\tilde{R})}\right)+g^{\frac{1}{2}}\left(\frac{\delta}{m-q}\right)\right)}{n\left(n^{\frac{\rho}{2}}-g^{\frac{1}{2}}\left(\frac{\delta}{m-q}\right)\right)},\notag
\end{align}
where
\begin{align}\label{equ:f_and_g_def}
    &f(\delta)\defeq\begin{cases}
        {2d\sigma^2\left(8\ln^{\frac{1}{2}}\left(\tfrac{4}{\delta}\right)+1\right)}\vspace{1mm} & {4\e^{-(nd)^2} \leq \delta \leq 1},\\[3mm]
        2\sigma^2\left(8\ln\left(\tfrac{4}{\delta}\right)+d\right) & {0 < \delta < 4\e^{-(nd)^2}},
    \end{cases}\\[1mm]
    &g(\delta) \,\defeq\, \ln\left(\tfrac{4d}{\delta}\right)2\hspace{0.3mm}\kappa\hspace{0.3mm} d.
\end{align}
}
\end{thm}
As expected, our bound for ridge regression is a bit more conservative than for the LS-based variant, however, it still decreases at an optimal rate of $\mathcal{O}(1/n)$. This additional conservatism is caused by the terms $2\lambda\ell^2d$ and $(1+2\lambda/\lambda_{\min}(\tilde{R}))$. Under Assumption \ref{assu:R_nonsigular}, $\lambda_{\min}(\tilde{R}) > 0$, and for nonvanishing excitation $1/\lambda_{\min}(\tilde{R})$ decreases as $n$ grows.
\begin{pf}
First, we investigate the case of one alternative sum, i.e., $m=2$, $q=1$, and give an upper bound on $r=\gamma^*=\gamma_1^*$ based on \citep{szentpeteri2025finite}. The solution of \eqref{equ:sps_eoa_cvx} only takes a bounded value $\gamma_0^* < \infty$ if and only if the RR-SPS region is bounded, hence
\begin{align}\label{equ:gammas_def}
    \gamma^* \defeq
    \begin{cases}
        \gamma_0^* &\text{if the RR-SPS region is bounded,}\\
        \infty & \text{if the RR-SPS region is unbounded.} \\
    \end{cases}
\end{align}
The SPS region is bounded iff both the perturbed and the unperturbed regressors span the whole space \citep{Care2022}. Consequently, under Assumption \ref{assu:R_nonsigular} and $m=2$, 
the RR-SPS region is bounded if and only if $A_1$ is positive definite.

Note that under Assumption \ref{assu:true_param}, a bounded confidence region can always be given, however, it does not influence the sample complexity, since the unbounded case is covered in the high probability bound by the complement event.

By using the thin QR-decomposition of $\Phi = \Phi_{{\scriptscriptstyle Q}}\Phi_{{\scriptscriptstyle R}}$ and introducing
\begin{equation}\label{equ:def_K}
    K \defeq \Phi_{{\scriptscriptstyle Q}}\tr D_{1}\Phi_{{\scriptscriptstyle Q}},
\end{equation}
we have that
\begin{equation}
    \xi \geq \frac{1}{1-\lambda_{\text{max}}(K^2)},
\end{equation}
and
\begin{align}\label{equ:gamma_ub}
    \gamma_0^* &= \xi^* \left(b_0\tr \left(A'_0(\xi^*)\right)^\dagger b_0 - c_0\right) \notag\\&
    \leq\frac{w\tr VD_dV\tr w(1+\norm{K}_2)}{n(1-\norm{K}_2)},
\end{align}
where $D_d \defeq \text{diag}(0,\dots,0,1,\dots,1)$ with $d$ ones and $V$ is orthonormal. As in the non-regularized case, for all $i$: $0<\abs{\lambda_i(K)} \leq 1$, and $K^2$ has an eigenvalue $1$, if and only if the region is unbounded. Hence, the upper bound for $\gamma_0^*$ (bounded case) can be extended to $\gamma$ (unbounded case) by considering $\gamma_0^* = $``$1/0$''$ = \infty$. Although the upper bound \eqref{equ:gamma_ub} for linear regression is proven through several intermediate lemmas \citep[Lemma 4, 5, 6, 7]{szentpeteri2025finite}, the extension of the matrices, see \eqref{equ:extended_matrices}, only affects Lemma 6, which also holds for $w_0 = w$ .

From the above and definition \eqref{equ:SPS_EOA_conf}, we obtain that 
\begin{align}
    \|\theta - \hat{\theta}\|^2_{\bar{R}}\leq\frac{w\tr VD_dV\tr w(1+\norm{K}_2)}{n(1-\norm{K}_2)}.
\end{align}
Next, we investigate the two terms, $w\tr VD_dV\tr w/n$ and $(1+\norm{K}_2)/(1-\norm{K}_2)$ separately. In order to do that three lemmas are introduced first. The first one is an improvement over \citep[Lemma 2]{szentpeteri2025finite} regarding its tightness.

\begin{lem}\label{lem:tilde_K_concentration}
    {\em
    Let\, $\tilde{\Phi} = U_{\tilde{\Phi}}\Sigma_{\tilde{\Phi}}V_{\tilde{\Phi}}\tr $ be the SVD decomposition of\, $\tilde{\Phi}$ and $\tilde{K} \defeq U_{\tilde{\Phi}}\tr\tilde{D}_1U_{\tilde{\Phi}}$. Under Assumption \ref{assu:tilde_Phi_U_coherence} we have with probability (w.p.) at least $1-\delta:$
    \begin{equation}
        \|\tilde{K}\|_2 \leq \left(\frac{2\kappa d\ln\left(\frac{2d}{\delta}\right)}{n^{\rho}}\right)^{\frac{1}{2}}.
    \end{equation}
    }
\end{lem}
The proof of Lemma \ref{lem:tilde_K_concentration} can be found in Section \ref{sec:tilde_K_concentration}. Building upon this result a concentration inequality for $(1+\norm{K}_2)/(1-\norm{K}_2)$ can be given. 
\begin{lem}\label{lemma:K_concentration}
    {\em
    Under Assumptions \ref{assu:R_nonsigular} and \ref{assu:tilde_Phi_U_coherence} we have with probability (w.p.) at least $1-\delta:$
    \begin{equation}
        \frac{1+\norm{K}_2}{1-\norm{K}_2}\leq\frac{1+\left(\frac{2\kappa d\ln\left(\frac{2d}{\delta}\right)}{n^{\rho}}\right)^{\frac{1}{2}} + \frac{2\lambda}{\lambda_{\min}(\tilde{R})}}{1-\left(\frac{2\kappa d\ln\left(\frac{2d}{\delta}\right)}{n^{\rho}}\right)^{\frac{1}{2}}}.
    \end{equation}
    }
\end{lem}
The proof of Lemma \ref{lemma:K_concentration} can be found in Section \ref{sec:K_concentration}.
The next and final lemma can be used to obtain a concentration inequality for $w\tr VD_dV\tr w/n$.

\begin{lem}\label{lemma:proj_indicator}
    {\em
    Under Assumption \ref{assu:noise} the following concentration inequality holds for 
    $w\tr  M w$, where $M$ is an orthogonal projection matrix with rank($M$) $\leq d$: w.p. at least $1-\delta:$
    \begin{align}
        &\frac{\vert w\tr  M w\vert}{n} \leq\\
        &\begin{cases}
            \dfrac{2d\sigma^2\left(8\ln^{\frac{1}{2}}(\tfrac{2}{\delta})+1\right)+2{\lambda}\norm {\theta^*}^2d}{n}\vspace{1mm} & {2\e^{-(nd)^2} \leq \delta \leq 1},\\[3mm]
            \dfrac{2\sigma^2\left(8\ln(\tfrac{2}{\delta})+d\right)+2{\lambda}\norm {\theta^*}^2d}{n} & {0 < \delta < 2\e^{-(nd)^2}},\notag
        \end{cases}
    \end{align}
    }
\end{lem}
The proof of Lemma \ref{lemma:proj_indicator} can be found in Section \ref{sec:proj_indicator}. 

Notice that $VD_dV\tr$ is a projection matrix with rank $d$, since $V$ is an orthonormal matrix and $D_d$ is a diagonal with $d$ values of 1 and $n-d$ values of 0 in its diagonal. Consequently, Lemma \ref{lemma:proj_indicator} can be applied to obtain w.p. at least $1-\delta/2$:
\begin{align}\label{equ:proj_result}
    &\frac{\vert w\tr  VD_dV\tr w\vert}{n} \leq\\
    &\begin{cases}
        \dfrac{2d\sigma^2\left(8\ln^{\frac{1}{2}}(\tfrac{4}{\delta})+1\right)+2{\lambda}\norm {\theta^*}^2d}{n}\vspace{1mm} & {4\e^{-(nd)^2} \leq \delta \leq 1},\\[3mm]
        \dfrac{2\sigma^2\left(8\ln(\tfrac{4}{\delta})+d\right)+2{\lambda}\norm {\theta^*}^2d}{n} & {0 < \delta < 4\e^{-(nd)^2}}.\notag
    \end{cases}
\end{align}

Combining \eqref{equ:proj_result}, Lemma \ref{lemma:K_concentration} with substituting $\delta/2$, and the union bound, i.e., if
\begin{align}\label{equ:union_1}
    &\BP(Y_1\leq y_1) \geq 1-p_1, &\BP(Y_2\leq y_2) \geq 1-p_2,
\end{align}
then
\begin{align}\label{equ:union_2}
    \BP(Y_1Y_2\leq y_1y_2) \geq 1-(p_1 + p_2),
\end{align}
we have w.p. at least $1-\delta$:
\begin{align}
    &\|\theta - \hat{\theta}\|^2_{\bar{R}} \leq \\&\dfrac{\left(f\left({\delta}\right)+2\lambda\norm {\theta^*}^2d\right)\left(n^{\frac{\rho}{2}}\left(1+\frac{2\lambda}{\lambda_{\min}(\tilde{R})}\right)+g^{\frac{1}{2}}\left({\delta}\right)\right)}{n\left(n^{\frac{\rho}{2}}-g^{\frac{1}{2}}\left({\delta}\right)\right)}\notag,
\end{align}
where $f(\delta)$ and $g(\delta)$ are defined in \eqref{equ:f_and_g_def}.
Using the upper bound of Assumption \ref{assu:true_param}, furthermore, the argument from the proof of \cite[Theorem 3]{szentpeteri2025finite} for arbitrary $m$ and $q$ values concludes the proof. \hfill $\qed$
\end{pf}
\subsection{Proof of Lemma \ref{lem:tilde_K_concentration}}\label{sec:tilde_K_concentration}
A concentration inequality on $\|\tilde{K}=
\sum_{t=1}^n \alpha_tu_{\tilde{\Phi},t}u_{\tilde{\Phi},t}\tr\|_2$, where $u_{\tilde{\Phi},t}$ are the row vectors of $U_{\tilde{\Phi}}$, can be given by building upon the ideas of \citep[Lemma 3]{szentpeteri2025} and \citep[Theorem 6.15]{wainwright_2019} as
\begin{align}\label{equ:prob_ub_sigmakappa_u_tilde_phi}
    &\BP\left(\norm{\sum_{t=1}^n \alpha_tu_{\tilde{\Phi},t}u_{\tilde{\Phi},t}\tr}_2\geq \varepsilon_0\right) \leq 2d\exp\left(-\frac{n\varepsilon_0^2}{2\sigma_{\kappa}^2}\right),
\end{align}
where
\begin{align}
    \sigma_{\kappa}^2
    &= \norm{\frac{1}{n}\sum_{t=1}^n \left(nu_{\tilde{\Phi},t}u_{\tilde{\Phi},t}\tr \right)^2}_2.
\end{align}
An improved upper bound on $\sigma_{\kappa}^2$ compared to \cite[Lemma 3]{szentpeteri2025} can be given by using the fact that $u_{\tilde{\Phi},t}$ is an orthonormal matrix
\begin{align}	
      \sigma_{\kappa}^2 &= n\norm{\sum_{t=1}^n 	\left(u_{\tilde{\Phi},t}u_{\tilde{\Phi},t}\tr \right)^2}_2  = n\norm{\sum_{t=1}^n 	\|{u_{\tilde{\Phi},t}}\|^2\left(u_{\tilde{\Phi},t}u_{\tilde{\Phi},t}\tr \right)}_2\notag\\
     &\leq n \max_{t \in [n]} \|{u_{\tilde{\Phi},t}}\|^2\norm{\sum_{t=1}^{n} u_{\tilde{\Phi},t}u_{\tilde{\Phi},t}\tr}_2 \notag\\
     &= n \max_{t \in [n]} \|{u_{\tilde{\Phi},t}}\|^2.
\end{align}
Using Assumption \ref{assu:tilde_Phi_U_coherence} 
\begin{align}
    \sigma_{\kappa}^2 \leq n \max_{t \in [n]} \|{u_{\tilde{\Phi},t}}\|^2\leq  n \, d\kappa n^{-\rho} = n^{1-\rho}d\kappa .
\end{align}
Substituting the upper bound on $\sigma_{\kappa}^2$ to \eqref{equ:prob_ub_sigmakappa_u_tilde_phi} and rewriting it to a stochastic lower bound completes the proof.\hfill $\qed$
\subsection{Proof of Lemma \ref{lemma:K_concentration}}\label{sec:K_concentration}
The QR decomposition of $\Phi$ is
\begin{equation}
    \Phi = \begin{bmatrix}
    \tilde{\Phi} \\
    \sqrt{{\lambda}}I
    % \sqrt{\tfrac{\lambda}{2}}I
\end{bmatrix} = \Phi_{{\scriptscriptstyle Q}}\Phi_{{\scriptscriptstyle R}},
\end{equation}
hence
\begin{equation}
    \Phi_{{\scriptscriptstyle Q}} = \begin{bmatrix}
    \tilde{\Phi}\Phi_{{\scriptscriptstyle R}}^{-1} \\
    % \sqrt{\tfrac{\lambda}{2}}\Phi_{{\scriptscriptstyle R}}^{-1}\\
    \sqrt{\lambda}\Phi_{{\scriptscriptstyle R}}^{-1}
\end{bmatrix},
\end{equation}
from which it follows that
\begin{align}
    K &= \Phi_{\scriptscriptstyle Q}\tr  D_1\Phi_{ \scriptscriptstyle Q} = \begin{bmatrix}
    \tilde{\Phi}\Phi_{{\scriptscriptstyle R}}^{-1} \\
    \sqrt{\lambda}\Phi_{{\scriptscriptstyle R}}^{-1}
    % \sqrt{\tfrac{\lambda}{2}}\Phi_{{\scriptscriptstyle R}}^{-1}
\end{bmatrix}\tr \begin{bmatrix}
        \tilde{D}_1 & 0\\
         0&I_d
         % 0&0&-I_d
    \end{bmatrix} \begin{bmatrix}
    \tilde{\Phi}\Phi_{{\scriptscriptstyle R}}^{-1} \\
    \sqrt{\lambda}\Phi_{{\scriptscriptstyle R}}^{-1}
    % \sqrt{\tfrac{\lambda}{2}}\Phi_{{\scriptscriptstyle R}}^{-1}
\end{bmatrix}\notag\\
    &= \Phi_{{\scriptscriptstyle R}}\ntr\tilde{\Phi}\tr\tilde{D}_1\tilde{\Phi}\Phi_{{\scriptscriptstyle R}}^{-1} + {\lambda}\Phi_{{\scriptscriptstyle R}}\ntr\Phi_{{\scriptscriptstyle R}}^{-1} 
\end{align}
Using the SVD decompositions $\tilde{\Phi} = U_{\tilde{\Phi}}\Sigma_{\tilde{\Phi}}V_{\tilde{\Phi}}\tr $, $\Phi_{R} = U_{\Phi_{R}}\Sigma_{\Phi_{R}}V_{\Phi_{R}}\tr $ and the facts that
\begin{align}\label{equ:R_reformulations}
    R & = \tilde{\Phi}\tr  \tilde{\Phi} + \lambda I = V_{\tilde{\Phi}}(\Sigma_{\tilde{\Phi}}^2 + \lambda I)V_{\tilde{\Phi}}\tr\notag\\
    R &= \Phi_{R}\tr \Phi_{R} =V_{\Phi_{R}}\Sigma_{\Phi_{R}}^2V_{\Phi_{R}}\tr,
\end{align}
we have that $(\Sigma_{\tilde{\Phi}}^2 + \lambda I) = \Sigma_{\Phi_{R}}^2$ and $V_{\Phi_{R}} = V_{\tilde{\Phi}}$ for the same arrangement of singular values. Consequently,
\begin{align}\label{equ:K_reformualtion}
    K &= U_{\Phi_{R}}\Sigma_{\Phi_{R}}^{-1}V_{\Phi_{R}}\tr V_{\Phi_{R}}\Sigma_{\tilde{\Phi}}U_{\tilde{\Phi}}\tr\tilde{D}_1U_{\tilde{\Phi}}\Sigma_{\tilde{\Phi}}V_{\Phi_{R}}\tr V_{\Phi_{R}}\Sigma_{\Phi_{R}}^{-1}U_{\Phi_{R}}\tr\notag\\
    &\quad +\lambda U_{\Phi_{R}}\Sigma_{\Phi_{R}}^{-1}V_{\Phi_{R}}\tr V_{\Phi_{R}}\Sigma_{\Phi_{R}}^{-1}U_{\Phi_{R}}\tr\notag\notag\\
    &= U_{\Phi_{R}}(\Sigma_{\tilde{\Phi}}^2 + \lambda I)^{-\frac{1}{2}}\Sigma_{\tilde{\Phi}}U_{\tilde{\Phi}}\tr\tilde{D}_1U_{\tilde{\Phi}}\Sigma_{\tilde{\Phi}}(\Sigma_{\tilde{\Phi}}^2 + \lambda I)^{-\frac{1}{2}}U_{\Phi_{R}}\tr\notag\\
    &\quad +\lambda U_{\Phi_{R}}(\Sigma_{\tilde{\Phi}}^2 + \lambda I)^{-1}U_{\Phi_{R}}\tr\notag\notag\\
    &=U_{\Phi_{R}}(\Lambda_1\tilde{K}\Lambda_1 + \Lambda_2)U_{\Phi_{R}}\tr,
\end{align}
where $\tilde{K} \defeq U_{\tilde{\Phi}}\tr\tilde{D}_1U_{\tilde{\Phi}}$, $\Lambda_1\defeq\Sigma_{\tilde{\Phi}}(\Sigma_{\tilde{\Phi}}^2 + \lambda I)^{-\frac{1}{2}}$ and $\Lambda_2\defeq\\\lambda(\Sigma_{\tilde{\Phi}}^2 + \lambda I)^{-1}$. Notice that as a consequence of \eqref{equ:R_reformulations}, we have $\Lambda_1 = \text{diag}((\lambda_i(\tilde{R})/(\lambda_i(\tilde{R}) + \lambda))^{1/2})_{i=1}^d$ and $\Lambda_2 = \text{diag}(\lambda/(\lambda_i(\tilde{R}) + \lambda))_{i=1}^d$, where $\lambda_i(\tilde{R})$ is the $i$-th eigenvalue of $\tilde{R}$. Furthermore, under Assumption \ref{assu:R_nonsigular}, it holds that $\Lambda_1^{-1}\Lambda_2\Lambda_1^{-1} = \lambda\Lambda_{\tilde{R}}^{-1}$, where $\Lambda_{\tilde{R}} = \text{diag}(\lambda_i(\tilde{R}))_{i=1}^d$.

We are aiming to give an upper bound for $\norm{K}_2$, therefore, we now investigate the eigenvalues of $K' \defeq \Lambda_1\tilde{K}\Lambda_1 + \Lambda_2 =\\ \Lambda_1(\tilde{K} + \lambda\Lambda_{\tilde{R}}^{-1})\Lambda_1$, see \eqref{equ:K_reformualtion}, using the Rayleigh quotient:
\begin{equation}
    R(K',x) \defeq \frac{x\tr\Lambda_1(\tilde{K} + \lambda\Lambda_{\tilde{R}}^{-1})\Lambda_1x}{x\tr x},
    \vspace{1mm}
\end{equation}
where $x$ is a non-zero vector. With a change of variables $y\defeq\Lambda_1 x$, where $y$ is also non-zero, we get
\begin{align}
    R(K',y) &=  \frac{y\tr(\tilde{K} + \lambda\Lambda_{\tilde{R}}^{-1})y}{y\tr \Lambda_1^{-2}y} = \frac{y\tr\tilde{K}y + y\tr\lambda\Lambda_{\tilde{R}}^{-1}y}{y\tr (I+\lambda\Lambda_{\tilde{R}}^{-1}) y}\\
    &=\frac{\frac{y\tr\tilde{K}y}{y\tr y} + \frac{y\tr\lambda\Lambda_{\tilde{R}}^{-1}y}{y\tr y}}{1+\frac{y\tr\lambda\Lambda_{\tilde{R}}^{-1}y}{y\tr y}} = \frac{R(\tilde{K}, y) + R(\lambda\Lambda_{\tilde{R}}^{-1}, y)}{1+R(\lambda\Lambda_{\tilde{R}}^{-1}, y)},\notag
\end{align}
where $\Lambda_1^{-2}=\Sigma_{\tilde{\Phi}}^{-2}(\Sigma_{\tilde{\Phi}}^2 + \lambda I)=I+\lambda\Lambda_{\tilde{R}}^{-1}$.
By introducing $\tilde{k}\defeq R(\tilde{K}, y)$ and $\tilde{r}\defeq R(\lambda\Lambda_{\tilde{R}}^{-1}, y)$ we can look at $R(K',y)$ as the function
\begin{equation}\label{equ:rayleigh_function}
    h(\tilde{k}, \tilde{r}) \defeq \frac{\tilde{k}+\tilde{r}}{1+\tilde{r}}.
\end{equation}
It holds for the partial derivatives of $h(\tilde{k}, \tilde{r})$ that
\begin{align}\label{equ:rayleigh_function_derivatives}
    \frac{\partial h}{\partial \tilde{k}} = \frac{1}{1+\tilde{r}} > 0,\qquad
    \frac{\partial h}{\partial \tilde{r}} = \frac{1-\tilde{k}}{(1+\tilde{r})^2} \geq 0,
\end{align}
where we used that $\tilde{k} \leq 1$, since by the property of the Rayleigh quotient, $\tilde{k} = R(\tilde{K}, y) \in [\lambda_{\min}(\tilde{K}), \lambda_{\max}(\tilde{K})]$ and $\|\tilde{K}\|_2 = \|U_{\tilde{\Phi}}\tr\tilde{D}_1U_{\tilde{\Phi}}\|_2\leq\|U_{\tilde{\Phi}}\tr\|\|\tilde{D}_1\|\|U_{\tilde{\Phi}}\|\leq 1$. As $h(\tilde{k}, \tilde{r})$ is strictly increasing in $\tilde{k}$ and non-decreasing in $\tilde{r}$, it takes its maximum, where $\tilde{k}$ and $\tilde{r}$ are maximized and its minimum, where they are minimized, therefore,
\begin{align}\label{equ:K_min_max_eigval}
    \lambda_{\max} (\Lambda_1\tilde{K}\Lambda_1 + \Lambda_2) &= \frac{\lambda_{\max}(\tilde{K}) + \lambda_{\max}(\lambda\Lambda_{\tilde{R}}^{-1})}{1+\lambda_{\max}(\lambda\Lambda_{\tilde{R}}^{-1})}\notag\\
        \lambda_{\min} (\Lambda_1\tilde{K}\Lambda_1 + \Lambda_2) &= \frac{\lambda_{\min}(\tilde{K}) + \lambda_{\min}(\lambda\Lambda_{\tilde{R}}^{-1})}{1+\lambda_{\min}(\lambda\Lambda_{\tilde{R}}^{-1})}
\end{align}
Now we can give an upper bound for $\|K\|_2$ using \eqref{equ:K_reformualtion}, \eqref{equ:K_min_max_eigval} and the triangle inequality as
\begin{align}
    &\|K\|_2 = \|U_{\Phi_{R}}(\Lambda_1\tilde{K}\Lambda_1 + \Lambda_2)U_{\Phi_{R}}\tr\|_2 = \|\Lambda_1\tilde{K}\Lambda_1 + \Lambda_2\|_2\notag\\
    &= \max\{\lambda_{\max} (\Lambda_1\tilde{K}\Lambda_1 + \Lambda_2), |\lambda_{\min} (\Lambda_1\tilde{K}\Lambda_1 + \Lambda_2)|\}\notag\\
    &=\max\Biggl\{\frac{\lambda_{\max}(\tilde{K}) + \frac{\lambda}{\lambda_{\min}(\tilde{R})}}{1+\frac{\lambda}{\lambda_{\min}(\tilde{R})}}, 
    \abs{\frac{\lambda_{\min}(\tilde{K}) + \frac{\lambda}{\lambda_{\max}(\tilde{R})}}{1+\frac{\lambda}{\lambda_{\max}(\tilde{R})}}}\Biggr\}\notag\\
    &\leq \max\Biggl\{\frac{\|\tilde{K}\|_2 + \frac{\lambda}{\lambda_{\min}(\tilde{R})}}{1+\frac{\lambda}{\lambda_{\min}(\tilde{R})}}, 
    {\frac{\|\tilde{K}\|_2 + \frac{\lambda}{\lambda_{\max}(\tilde{R})}}{1+\frac{\lambda}{\lambda_{\max}(\tilde{R})}}}\Biggr\}.
\end{align}
Using the same ideas as in \eqref{equ:rayleigh_function}-\eqref{equ:rayleigh_function_derivatives}, i.e., these are monotone functions and it takes the maximum when it is $\lambda_{\min}(\tilde{R})$, we obtain
\begin{equation}
    \|K\|_2 \leq \frac{\|\tilde{K}\|_2 + \frac{\lambda}{\lambda_{\min}(\tilde{R})}}{1+\frac{\lambda}{\lambda_{\min}(\tilde{R})}},
\end{equation}
therefore
\begin{equation}
    \frac{1+\norm{K}_2}{1-\norm{K}_2}\leq\frac{1+\|{\tilde{K}}\|_2 + \frac{2\lambda}{\lambda_{\min}(\tilde{R})}}{1-\|{\tilde{K}}\|_2}.
\end{equation}
Applying Lemma \ref{lem:tilde_K_concentration} completes the proof.
\hfill$\qed$
\subsection{Proof of Lemma \ref{lemma:proj_indicator}}\label{sec:proj_indicator}
    The projection matrix $M$ can be factorized using the eigendecomposition as
    \begin{align}
        w\tr  M w = w\tr  V_{\scriptscriptstyle M} \Lambda_{\scriptscriptstyle M} V_{\scriptscriptstyle M}\tr  w.
    \end{align}
    Let $d'\defeq$ rank($M$), then, the projection matrix $M$ has $d'$ eigenvalues of 1 and $n-d'$ eigenvalues of 0, therefore $\Lambda_{\scriptscriptstyle M} = D_{d'} = \text{diag}(0,\dots,0,1,\dots,1)$. By defining $w'\defeq D_{d'} V_{\scriptscriptstyle M}\tr w$ we have that $w' \in \mathbb{R}^{d'}$ and
    \begin{align}\label{equ:wMw_as_norm}
        \norm{w'}^2 = \norm{D_{d'} V_{\scriptscriptstyle M}\tr  w}^2 = w\tr  M w.
    \end{align}
    Every component of $w'$ can be written as $w'_{i} = w\tr  v_{{\scriptscriptstyle M}, i}$, where $v_{{\scriptscriptstyle M}, i}$ is an eigenvector of $M$ corresponding to a non-zero eigenvalue. Using \eqref{equ:extended_matrices} and vector notation $v_{i:j}\defeq [v_i,\dots,v_j]\tr$, we obtain 
    \begin{align}
        w'_{i} &= \tilde{w}\tr  v_{{\scriptscriptstyle M}, i, 1:n} -\sqrt{\lambda}(\theta^*)\tr v_{{\scriptscriptstyle M}, i, n+1:n+d}.
    \end{align}
    The norm $\norm{w'}^2$ can be upper bounded as 
    \begin{align}\label{equ:wMw_ub}
        \norm{w'}^2 &= \sum_{i=1}^{d'} ({w'}_i)^2 \notag\\
        &\leq 2\sum_{i=1}^{d'} (\tilde{w}\tr  v_{{\scriptscriptstyle M}, i, 1:n})^2+{\lambda}((\theta^*)\tr v_{{\scriptscriptstyle M}, i, n+1:n+d})^2\notag\\
        &\leq 2\sum_{i=1}^{d'} (\tilde{w}\tr  v_{{\scriptscriptstyle M}, i, 1:n})^2  + {\lambda}\norm {\theta^*}^2\norm{v_{{\scriptscriptstyle M}, i, n+1:n+d}}^2\notag\\
        &\leq 2{\lambda}\norm {\theta^*}^2d + 2\sum_{i=1}^{d} (\tilde{w}\tr  v_{{\scriptscriptstyle M}, i, 1:n})^2,
    \end{align}
    where we used our assumption $d'\leq d$, the Cauchy–Schwarz inequality in the second and third line and the fact that $v_{{\scriptscriptstyle M}, i}$ is an eigenvector in the last line.
    Using the results of \cite[Lemma 2]{szentpeteri2025} and its proof,
    it follows that $X \defeq \sum_{i=1}^{d} (\tilde{w}\tr  v_{{\scriptscriptstyle M}, i, 1:n})^2$ is subexponential with parameters $(4d \sigma^2\sqrt{2},4\sigma^2)$.
    Then, the following inequality holds for $X$ \cite[Proposition 2.9]{wainwright_2019}:
    \begin{align}\label{equ:X_concentration}
        &\BP\left(\frac{\vert X - \mathbb{E}X\vert}{n}\hspace{-0.3mm} \geq\hspace{-0.3mm} \varepsilon\right) \hspace{-0.3mm}\leq\hspace{-0.3mm}
        \begin{cases}
            2\exp(-\frac{\varepsilon^2n^2}{64d^2\sigma^4}) &\! 0 \leq \varepsilon \leq 8\sigma^2 {d^2},\\[2mm]
            2\exp(-\frac{\varepsilon n}{8\sigma^2}) &\! \varepsilon > 8\sigma^2{d^2}.
        \end{cases}
    \end{align}
    Also, applying the same arguments as in \cite[Lemma 4]{szentpeteri2025}, inequality \eqref{equ:X_concentration} can be reformulated as, w.p. at least $1-\delta$:
    \begin{align}\label{equ:X_concentration_lower}
        &\frac{\vert X\vert}{n} \leq
        \begin{cases}
            \dfrac{d\sigma^2\left(8\ln^{\frac{1}{2}}(\tfrac{2}{\delta})+1\right)}{n}\vspace{1mm} & {2\e^{-(nd)^2} \leq \delta \leq 1},\\[3mm]
            \dfrac{\sigma^2\left(8\ln(\tfrac{2}{\delta})+d\right)}{n} & {0 < \delta < 2\e^{-(nd)^2}}.
        \end{cases}
    \end{align}
    Combining \eqref{equ:wMw_as_norm}, \eqref{equ:wMw_ub} and \eqref{equ:X_concentration_lower} we have, w.p. at least $1-\delta$:
      \vspace{-2mm}
      %\begin{equation}
      %\begin{aligned}
        \begin{align}
        &\frac{\vert w\tr  M w\vert}{n} \leq\\
        &\begin{cases}
            \dfrac{2d\sigma^2\left(8\ln^{\frac{1}{2}}(\tfrac{2}{\delta})+1\right)+2{\lambda}\norm {\theta^*}^2d}{n}\vspace{1mm} & {2\e^{-(nd)^2} \leq \delta \leq 1},\\[3mm]
            \dfrac{2\sigma^2\left(8\ln(\tfrac{2}{\delta})+d\right)+2{\lambda}\norm {\theta^*}^2d}{n} & {0 < \delta < 2\e^{-(nd)^2}},\notag
        \end{cases}\\[-4mm]\notag
        \end{align}
    %\end{aligned}
    %\end{equation}
    which concludes the proof. \hfill$\qed$

\section{Simulation Experiments}

In this section we illustrate how regularization affects the size of RR-SPS EOA regions in a series of numerical experiments. We consider a $2$-dimensional FIR system:
\begin{align}\label{equ:fir_exp}
    Y_t = b_1^*U_{t-1} + b_2^*U_{t-2} + W_t,
\end{align}
where $b_1^*=b_2^*=2$, and the input is given by 
\begin{equation}
    U_t = aU_{t-1} + \sum_{i=1}^3c_iV_{t-i+1},
\end{equation}
with $a=0.7$, $c_1=0.9$, $c_2=0.5$, $c_3=0.1$, and $V_t$ follows a standard normal distribution $\mathcal{N}(0, 1)$.

\begin{figure}[t!]
\begin{center}
\includegraphics[width=8.4cm]{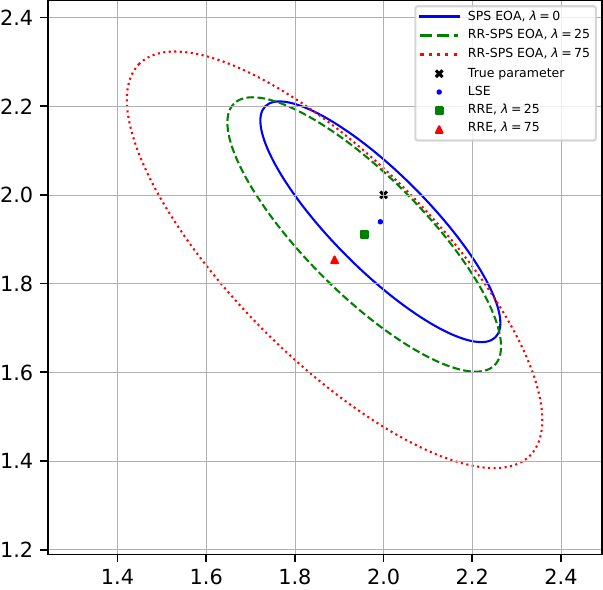}    % The printed column width is 8.4 cm.
\caption{Sizes of $0.9$-level RR-SPS EOA regions with different regularization parameter values $\lambda$ for $n=250$.} 
\label{fig:reg_sps_diff_lambda}
\end{center}
\end{figure}

In the first experiment we investigated how a large regularization parameter influences the size of the RR-SPS EOA regions, hence we set $\lambda_1 = 25$ and $\lambda_2 = 75$. The sample size was $n=250$, $\{W_t\}$ were i.i.d. standard Laplacian random variables and we built $0.9$-level confidence regions with $m=10$ and $q=1$. Our result is illustrated in Fig. \ref{fig:reg_sps_diff_lambda}. 

It can be observed that increasing $\lambda$ leads to a more biased estimate, hence, to larger confidence regions, since they contain the true parameter with (at least) a given probability. This dependence on $\lambda$ and $\|\theta^*\|$ appears in our upper bound on the size of the regions (Theorem \ref{thm:sample_complex_eoa}). 

In a second experiment, we compared the empirical size of $0.9$-level ($m=10$, $q=1$) RR-SPS EOA regions with our theoretical bound in Theorem \ref{thm:sample_complex_eoa}, furthermore, with the sizes of non-regularized SPS EOA 
(incorporating the improvements of Lemma \ref{lem:tilde_K_concentration})
and asymptotic confidence regions. The same 
setting was used as in the first experiment, with the exception that $\{W_t\}$ were i.i.d. variables distributed uniformly
on $(-2, 2)$. It follows that the optimal subgaussian variance proxy was $\sigma^2={16/12}$. We used $\rho=1$, $\lambda=10$, where regularization was applied, and set $\kappa$ to the empirical value $\frac{n}{d} \max_{1\leq i \leq n}\|U_{\tilde{\Phi}}\tr e_i\|^2$, 
for every $n$. The median of the empirical sizes from $s=100$ independent 
simulations and the theoretical bounds 
with $\delta=0.5$ are presented in Table \ref{tab:conf_size_compare}. for different sample sizes.

The result shows that our theoretical bound on the RR-SPS EOA sizes matches the decrease rate of the empirical sizes, however, for smaller sample sizes, they tend to be conservative, especially for a large $\lambda$ and $\|\theta^*\|$. This is not surprising, since our bound applies to a broad class of data realizations for the chosen hyperparameters, whereas the RR-SPS EOA builds data-driven confidence ellipsoids. 

{
\renewcommand*{\arraystretch}{1.2}
\begin{table}[t!]
\begin{center}
\caption{Comparison of $0.9$-level confidence regions with $\lambda=10$, $s=100$, $\delta=0.5$.}\label{tab:conf_size_compare}
\begin{tabular}{lccccc}
sample size ($n$) & 250 & 500 & 1000 & 1500 & 2000 \\\hline
emp.\ SPS EOA & 0.038 & 0.017 & 0.007 & 0.005 & 0.004 \\
emp.\ RR-SPS EOA & 0.042 & 0.019 & 0.008 & 0.006 & 0.004 \\
th.\ SPS EOA & 4.931 & 0.895 & 0.254 & 0.136 & 0.091 \\
th.\ RR-SPS EOA & 24.03 & 4.168 & 1.158 & 0.615 & 0.409 \\
emp.\ asymptotic & 0.025 & 0.012 & 0.006 & 0.004 & 0.003 \\
\hline
\end{tabular}
\end{center}
\end{table}
}

\section{Conclusion}
In this paper, we introduced a generalization of the SPS EOA confidence ellipsoid construction method to ridge regression. We analyzed the size of the resulting regions by deriving PAC-based upper bounds and showed how regularization influence their magnitude without altering the optimal decrease rate. Due to regularization and new technical insights we achieved tighter bounds under weaker assumptions compared to the previous results.

Further research directions include studying how regularization affects the SPS regions for feedback systems and applying our result to stochastic linear bandits.

\bibliography{ifacconf}
                                                   
\end{document}